\documentclass[lettersize,journal]{IEEEtran}

\usepackage{graphicx, color}  

\usepackage{algorithm}  
\usepackage{algpseudocode} 

\usepackage{bm}
\usepackage{amsmath, amsfonts}

\usepackage{cite}
\usepackage{url}
\usepackage{enumerate}

\usepackage[export]{adjustbox}
\usepackage{subfig}
\usepackage{afterpage}
\usepackage{multirow}
\usepackage[procnames]{listings}

\definecolor{keywords}{RGB}{255,0,90}
\definecolor{red}{RGB}{160,0,0}
\definecolor{green}{RGB}{0,150,0}

\usepackage{amsthm}
\newtheoremstyle{indented}{3pt}{3pt}{}{}{\bfseries}{.}{.5em}{}
\theoremstyle{indented}

\begin{document}
	
	\title{Inherently Interpretable Tree Ensemble Learning}
	\author{Zebin Yang,
		Agus Sudjianto,  
		Xiaoming Li,
		and Aijun Zhang~\IEEEmembership{Member,~IEEE}
		\IEEEcompsocitemizethanks{
			\IEEEcompsocthanksitem Zebin Yang, Xiaoming Li, and Aijun Zhang are with Wells Fargo, USA. E-mail: \{Zebin.Yang, Xiaoming.Li, Aijun.Zhang\}@wellsfargo.com. Agus Sudjianto is with H2O.ai and University of North Carolina at Charlotte. E-mail: agus.sudjianto@h2o.ai. Aijun Zhang is the corresponding author.
		\IEEEcompsocthanksitem The views expressed in this paper are those of the authors and do not necessarily reflect those of Wells Fargo.}
	}
 
	\markboth{IEEE Transactions on on Neural Networks and Learning Systems}
	{Yang \MakeLowercase{\textit{et al.}}: Inherently Interpretable Tree Ensemble Learning}
	
	\IEEEtitleabstractindextext{
		\begin{abstract}
			Tree ensemble models like random forests and gradient boosting machines are widely used in machine learning due to their excellent predictive performance. However, a high-performance ensemble consisting of a large number of decision trees lacks sufficient transparency and explainability. In this paper, we demonstrate that when shallow decision trees are used as base learners, the ensemble learning algorithms can not only become inherently interpretable subject to an equivalent representation as the generalized additive models but also sometimes lead to better generalization performance. First, an interpretation algorithm is developed that converts the tree ensemble into the functional ANOVA representation with inherent interpretability. Second, two strategies are proposed to further enhance the model interpretability, i.e., by adding constraints in the model training stage and post-hoc effect pruning. Experiments on simulations and real-world datasets show that our proposed methods offer a better trade-off between model interpretation and predictive performance, compared with its counterpart benchmarks.
		\end{abstract}
		}
	
	\maketitle
	
	\IEEEdisplaynontitleabstractindextext
	\IEEEpeerreviewmaketitle

	\section{Introduction}
	\IEEEPARstart{T}ree ensembles are widely recognized as one of the most popular machine learning techniques for modeling tabular data. For example, a bagging tree aggregates multiple regression or classification trees by making bootstrap replicates of the training data~\cite{breiman1996bagging}. The random forest also averages a bunch of decision trees to reduce the variance, and it combines the bagging and random feature selection strategies to draw training samples for every single tree~\cite{breiman2001random}. In contrast to constructing trees independently, gradient-boosted machines employ a sequential fitting approach. Each new tree in the ensemble is added to address the deficiencies of the previous trees and enhance the model's performance~\cite{friedman2001greedy}.
	
	In general, gradient-boosting trees tend to exhibit superior predictive performance compared to random forests and bagging trees. The state-of-the-art implementations of gradient-boosted machines include XGBoost~\cite{chen2016xgboost}, LightGBM~\cite{ke2017lightgbm}, and CatBoost~\cite{dorogush2018catboost}, in which they have developed a wide range of extensions and enhancements built upon the na\"ive algorithm. Although tree ensemble models demonstrate superb predictive performance, they often suffer from the model interpretation challenge. A well-performing tree ensemble model usually consists of a large number of trees. Each tree can be interpreted separately, but it becomes almost impossible to understand and interpret the whole model. As a result, tree ensemble models are usually perceived as black boxes.
	
	Functional ANOVA~\cite{stone1994use,huang1998projection} is a promising framework for interpreting black-box models. It decomposes a model as the sum of additive components, as follows.
	\begin{align}
		f(\textbf{x})  = \mu + \sum\limits_{j} f_{j}(x_{j}) + \sum\limits_{jk} f_{jk}(x_{j},x_{k}) + \ldots,  \label{fanova}
	\end{align}
	where $\mu$ is the global intercept, $f_{j}(x_{j})$ is the main effect, $f_{jk}(x_{j}, x_{k})$ is the pairwise interaction, $f_{jkl}(x_{j} ,x_{k}, x_{l})$ is the 3-way interaction, etc. In a tree ensemble, the highest interaction order is upper-bounded by the number of features and the maximum tree depth. In this paper, we demonstrate that when the shallow decision trees are used as base learners, tree ensemble models can not only become inherently interpretable but also sometimes lead to better generalization performance. Our main contributions in this paper are summarized below.
	\begin{itemize}
		\item Develop a functional ANOVA-based algorithm to interpret tree ensemble models. This interpretation is faithful to the original model, and the decomposed main effects and interactions can be easily explained.
		\item Propose two strategies to further enhance the interpretability of tree ensemble models: a) adding hard constraints (e.g., maximum depth and monotonicity constraints) or soft regularization (e.g., L1 and L2 regularization); and b) pruning trivial effects after functional ANOVA transformation, to make the model parsimonious. With these enhancements, a better trade-off between model interpretability and predictive performance can be achieved.
		\item Showcase the proposed methods through simulation and real datasets and compare them with benchmark inherently interpretable methods. Compared with the state-of-the-art explainable boosting machine (EBM;~\cite{lou2013accurate}), tree ensemble models with a depth of 2 can achieve better predictive performance and interpretability. If necessary, we can also add high-order interactions to the model.
	\end{itemize}
	
	It is our hope to promote the use of shallow tree ensemble models for tabular data modeling. The rest of the paper is organized as follows. Section~\ref{related_word} reviews related literature. In Section~\ref{methodology}, the proposed functional ANOVA-based interpretation algorithm is introduced. Following that, Section~\ref{enhancement} provides several interpretability enhancement strategies for practical applications. The numerical experiments are shown in Section~\ref{experiments}. Finally, Section~\ref{conclusion} concludes this paper.
	
	\section{Related Work} \label{related_word}
	The techniques in interpretable machine learning can be roughly classified into post-hoc explanation tools and inherently interpretable models. The former aims at explaining an arbitrary fitted model, and it can be further divided into global and local explanations. Examples of global explanation include partial dependence plot (PDP;~\cite{friedman2001greedy}) and accumulative local effects (ALE;~\cite{apley2020visualizing}), where both of them are used to reveal the relationship between one or two features and the model prediction. In contrast, local explanation methods like the local interpretable model-agnostic explanation (LIME;~\cite{ribeiro2016should}) and Shapley additive explanations (SHAP;~\cite{lundberg2017unified,lundberg2020local}) decompose the prediction outcome of an individual sample into the contributions of each feature. The primary drawback of post-hoc explanation tools is that the interpretation results are mere approximations, which may deviate from the original model and be incorrect or unfaithful~\cite{rudin2019stop}. This can pose significant risks, especially in sensitive domains such as healthcare and finance.
	
	The second category aims at developing inherently interpretable models. This is in contrast to black-box models (e.g., neural networks), in which the decision-making process is too complicated to interpret. In practice, much of the complexity is unnecessary and may lead to overfitting. The key idea of inherently interpretable models is to regularize or constrain complex models to be interpretable, without sacrificing the predictive performance. Some principles of designing interpretable models include additivity, sparsity, smoothness, etc~\cite{sudjianto2021designing}. For instance, an explainable boosting machine (EBM;~\cite{lou2013accurate}) is a generalized additive model with functional pairwise interactions. It fits the main effects and interactions sequentially using shallow tree ensemble models. The generalized additive model with structured pairwise interactions network (GAMI-Net;~\cite{yang2021gami}) is an alternative to EBM, but uses modularized neural networks to estimate the main effects and pairwise interactions. The GAMI-Lin-T~\cite{hu2023interpretable} model is another recently proposed interpretable model under the functional ANOVA framework. It also uses the boosting algorithm and the base learners are trees with linear functions in leaves.
	
	The proposed interpretation algorithm combines elements from both categories mentioned above, serving as a post-hoc tool specifically for interpreting tree ensemble models. Notably, it endows tree ensemble models with inherent interpretability, ensuring the derived interpretations are precise without any approximation. Additionally, a recently introduced effect purification algorithm~\cite{lengerich2020purifying} is incorporated to tackle the identifiability problem between main effects and their corresponding interaction effects under the functional ANOVA framework. This paper leverages this purification algorithm to convert tree ensemble models into a functional ANOVA-based representation.
	
	In the literature, there exist some attempts to interpret shallow tree ensemble models. For example, the decision stump boosting~\cite{oliver1994averaging,denison2001boosting} uses decision trees with only one split as base learners, and the resulting model can be represented as a generalized additive model. The EBM models share the same model form as tree ensemble models as the maximum tree depth is 2. Both of them are composed of main effects and pairwise interactions, and the effect functions are piecewise constant. The main difference is in the model fitting method. In EBM, the main effects are fitted first in a round-robin fashion, and followed by the pairwise interactions. In contrast, tree ensemble models fit all effects greedily without any predefined order, and therefore, tend to have better predictive performance.

	\section{Interpret Tree Ensemble Models} \label{methodology}
	A tree ensemble model can be represented as the addition of (tree, weight)-pairs
	\begin{equation}
		f(\textbf{x})=\sum_{k=1}^K w_k T_k(\boldsymbol{x}), \label{tree_ensemble}
	\end{equation}
	where $K$ is the total number of trees and $g$ is a link function, which is an identity function for regression or a logit function for binary classification. In gradient boosting, the weights $w_k$ correspond to the learning rates. Each tree $T_k$ can be further represented as the addition of leaf nodes. By rearranging the additive components, we can represent (\ref{tree_ensemble}) as the addition of all leaf nodes, as follows, 
	\begin{equation}
		f(\textbf{x})=\sum_{m=1}^{M} v_{m} \prod_{j \in S_{m}} I\left(s_{mj}^{l} \leq x_{j} < s_{mj}^{u}\right), \label{leaf_ensemble}
	\end{equation}
	where $M$ is the total number of leaf nodes and $v_{m}$ is the value of the $m$-th leaf node, multiplied by the corresponding tree weight. The symbol $S_{m}$ represents the set of split variables in the decision path of the $m$-th leaf node. The product of indicator functions denotes whether a sample belongs to the corresponding leaf node. In specific, the interval $[s_{mj}^{l}, s_{mj}^{u})$ is determined by the following rules.
	
	\begin{itemize}	
		\item If a tree has no split, then $s_{mj}^{l}=-\inf$ and $s_{mj}^{u}=\inf$. This is a special case where the root node stops splitting, and it corresponds to an intercept term.
		
		\item As a feature is used only once in the decision path, and the leaf node belongs to the left side of the split point $s$, then $s_{mj}^{u}=s$ and $s_{mj}^{l}=-\inf$. Otherwise, if the leaf node belongs to the right side of the split point, then $s_{mj}^{l}=s$ and $s_{mj}^{u}=\inf$.
		
		\item As a feature is used multiple times in the decision path, then $s_{mj}^{l}$ and $s_{mj}^{u}$ are determined by the intersection of these split-generated intervals.
	\end{itemize}
	
	Given the basic representation of tree ensemble models, we can further interpret it using the functional ANOVA framework. The algorithm can be divided into three steps, i.e., aggregation, purification, and attribution.	Note that the proposed interpretation method applies to the bagging of multiple tree-ensemble models, as each additive ensemble is also inherently interpretable.
	
	\subsection{Aggregation}
	The first step is to rearrange (\ref{leaf_ensemble}) using the functional ANOVA framework defined in (\ref{fanova}), by assigning each leaf node to the effect functions. For each leaf node, its corresponding effect function is determined by the distinct split variables at its decision path. For example, leaf nodes with only one distinct split variable are the main effects. The $j$-th main effect $f_{j}(x_{j})$ is obtained by the sum of all the leaf nodes functions subject to $S_{m}=\{j\}$, as follows,
	\begin{equation}
		f_{j}(x_{j})=\sum_{S_{m} = \{j\}} v_{m} \cdot I\left(s_{mj}^{l} \leq x_{j} < s_{mj}^{u}\right). \label{leaf_maineffects}
	\end{equation}
	Leaf nodes with two distinct split variables correspond to pairwise interactions. A pairwise interaction $f_{jk}(x_{j}, x_{k})$ can be calculated by the sum of all the leaf nodes subject to $S_{m}=\{j, k\}$, as follows,
	\begin{equation}
		\begin{aligned}
		f_{jk}(x_{j}, x_{k})= & \sum_{S_{m} = \{j,k\}} v_{m} \cdot I\left(s_{mj}^{l} \leq x_{j} < s_{mj}^{u}\right) \\
		&
		\cdot I\left(s_{mk}^{l} \leq x_{k} < s_{mk}^{u}\right). \label{leaf_interactions}
		\end{aligned}
	\end{equation}
	
	Similarly, leaf nodes with more than two distinct split variables are assigned to the corresponding higher-order interaction terms. For a depth-$d$ tree ensemble model, each leaf node would have at most $d$ distinct split variables, and hence the highest possible interaction order is also $d$. In particular, a shallow tree ensemble with a maximum depth of 1 can be represented as a generalized additive model (GAM). A depth-2 tree ensemble can be represented as a generalized additive model with pairwise interaction (GAMI), etc. 
	
	Note that all effect functions are piece-wise constant representing a weighted sum of indicator functions. A main effect $f_{j}(x_{j})$ with $N_j$ distinct split points can be represented as a value vector of length $N_j+1$. A pairwise interaction $f_{jk}(x_{j}, x_{k})$ with $N_j$ and $N_k$ distinct split points on features $j$ and $k$, respectively, can be represented as a matrix of size $(N_j + 1, N_k + 1)$. In general, higher-order effects can also be represented using higher-order tensors, using a similar approach.
	
	\subsection{Purification}
	The functional ANOVA would suffer from the identifiability issue without any constraint. For example, a main effect term can be absorbed into its parent interactions without changing the model prediction. This will lead to multiple equivalent representations and make the interpretation non-unique. To ensure a unique interpretation, it is assumed that the decomposed effects satisfy the following constraint
	\begin{equation}
		\int f_{i_1 \cdots i_t}\left(x_{i_1}, \cdots, x_{i_t}\right) d x_k=0, \quad k=i_1, \cdots, i_t, \label{fanova_constraint}
	\end{equation}
	where $i_1, \cdots, i_t$ are feature indices. It implies that all main and interaction effects a) have zero means and b) are mutually orthogonal, i.e.,
	\begin{equation}
		\int f_{i_1 \cdots i_u}\left(x_{i_1}, \cdots, x_{i_u}\right) f_{j_1 \cdots j_v}\left(x_{j_1}, \cdots, x_{j_v}\right) d \textbf{x} =0,
	\end{equation}
	whenever $\left(i_1, \cdots, i_u\right) \neq\left(j_1, \cdots, j_v\right)$.
	
	In the aggregation step, we have rearranged all the leaf node rules to the corresponding effects. However, these raw effects do not necessarily satisfy the functional ANOVA constraint in (\ref{fanova_constraint}). To address this issue, an effect purification algorithm proposed by \cite{lengerich2020purifying} is applied. For an arbitrary effect $f_{i_1 \cdots i_t}\left(x_{i_1}, \cdots, x_{i_t}\right)$, it approximates (\ref{fanova_constraint}) by removing the means of each slice feature $i_1 \cdots i_t$ iteratively and sequentially. The removed effects are then added to the corresponding child effects, to ensure the equivalence of the purified model and the original model.
	
	For simplicity, we illustrate this algorithm using a pairwise interaction $f_{jk}(x_{j}, x_{k})$. We take the matrix representation of the pairwise interaction (of size $(N_j + 1, N_k + 1)$) as input. This algorithm then operates the matrix using the following steps:
	\begin{itemize}
		\item Calculate the average value along the first dimension, and get a mean vector of size $(N_k + 1)$. Subtract the mean vector from the value matrix, and add it to the corresponding main effect $f_{k}(x_{k})$.
		
		\item Calculate the average value along the second dimension, and get the mean vector of size $(N_j + 1)$. Subtract the mean vector from the value matrix, and add it to the corresponding main effect $f_{j}(x_{j})$.
	\end{itemize}
	These two steps are repeated multiple times until convergence, i.e., as the absolute difference of the matrix between two consecutive iterations is less than a predefined threshold. In the end, we would get a purified pairwise interaction, as well as two updated child main effects. In general, for a $d$-way interaction, the purification algorithm would iterate over each dimension of the corresponding $d$-way tensor, and for each dimension, it moves the $(d-1)$-way mean tensor to the corresponding child $(d-1)$-way interaction. The final result would be a purified $d$-way interaction, together with $d$ child $(d-1)$-way interactions. 
	
	The whole purification algorithm would start from the highest-order interactions and recursively cascade effects from high-order interactions to low-order interactions. Finally, for main effects, we can simply center them to have zero means, and the subtracted mean is then added to the intercept term. As the purification step finishes, we can visualize the main effects through 1D line plots and pairwise interactions via 2D heatmaps. For higher-order interactions, we can draw 1D or 2D plots for one or two features of interest, while fixing the rest features to certain representative values.
	
	In the above discussion, we assume the data is uniformly and independently distributed over the feature space, which may be not the case in practical applications. The weighted functional ANOVA decomposition~\cite{hooker2007generalized} is accordingly proposed by considering the empirical distribution of data. To use weighted functional ANOVA, we first calculate the density for each bin of the matrix / tensor, and the simple average is replaced by the weighted average.
	
	\subsection{Attribution}
	As we have converted a tree ensemble model into the functional ANOVA representation, the next step is to quantify the contribution of the decomposed effects, both locally (for an individual sample) and globally (for the entire dataset). Below we introduce the definition of effect contributions and feature contributions. 
	
	The effect-level contribution quantifies the contribution of each effect to the model output. For example, the contribution of the $j$-th main effect is $f_{j}(x_j)$, and $f_{jk}(x_j, x_k)$ is the contribution of the pairwise interaction $(j, k)$, etc. 
	\begin{itemize}
		\item \textbf{Local effect contribution}: The model output for each sample can be interpreted as the sum of all effect contributions plus the intercept term. Each effect contribution can have a positive, negative, or zero value. By considering the magnitude of effect values, we can select the most significant effects for an individual sample.
		\item \textbf{Global effect importance}: After calculating the effect contributions for each sample, we can summarize the importance of each effect by examining the variance of the local effect contributions across a given dataset, such as the training data. Subsequently, the effect importance is normalized in a way that ensures the sum of all effects' importance equals 1.
	\end{itemize}
	
	In contrast, the feature-level contribution quantifies the contribution of a feature $j$ to the model output of an individual sample, i.e.,
	\begin{equation}
	\begin{aligned}
			z_j(x_j)  = & f_{j}(x_{j}) + \frac{1}{2} \sum\limits_{jk} f_{jk}(x_{j}, x_{k}) +                                                                              \\
			& \frac{1}{3} \sum\limits_{jkl} f_{jkl}(x_{j}, x_{k}, x_{l}) + \cdots + \frac{1}{p}f_{1 \cdots p}(x_1, \cdots, x_p). \label{feature_contribution}
	\end{aligned}
	\end{equation}

	In this formula, the main effect $f_j(x_j)$ is added directly to the $j$-th feature contribution. Additionally, all pairwise interaction effects associated with feature $j$ are included in the feature contribution, but with a discount factor of 2. This rule is also extended to 3-way, 4-way, and up to $p$-way interactions, where $p$ is the number of features.
	\begin{itemize}
		\item \textbf{Local feature contribution}: Similar to the local effect contribution, we can locally interpret the model output of an individual sample at the feature level, i.e., by $	z_j(x_j)$.
		\item \textbf{Global feature importance}: The significance of feature $j$ is determined by evaluating the variance of $z_j(x_j)$ on a specific dataset, such as the training data. After that, we normalize the feature importance to ensure that the total importance of all features adds up to 1.
	\end{itemize}
	
	It is important to note that the feature contribution $z_j(x_j)$ is exactly the Shapley value~\cite{shapley1953value} of feature $j$, defined as follows,
	\begin{equation}
		\phi_j=\sum_{S \subseteq\{1, \ldots, p\} \backslash \{j\}} \frac{|S| !(p-|S|-1) !}{p !}\left(v(S \cup\{j\})-v(S)\right), \label{shapley_value}
	\end{equation}
	where $v$ is the value function that returns the prediction of each feature coalition $S$. The marginal contribution of feature $j$ to the coalition $S$ is quantified by $v(S \cup \{j\}) - v(S)$, and the multiplier on the left is the weight of feature coalitions. In the functional ANOVA framework, the value function of different feature coalitions is already defined. The proof of the equivalence between Shapley value and feature contribution $z_j(x_j)$ can be found in \cite{owen2014sobol}. In shallow tree ensemble models, we can exactly calculate the Shapley value / feature contribution without much computational burden.

	\section{Interpretability Enhancement} \label{enhancement}
	In this section, we discuss several strategies to enhance the interpretability of tree ensemble models.
	
	\subsection{Interpretability Constraints}
	A tree ensemble model can be made more interpretable by adjusting its hyperparameters. Here we introduce some of the most important hyperparameters in the state-of-the-art tree ensemble learning frameworks.
	
	\subsubsection{Maximum Tree Depth} 
	In the full functional ANOVA representation, the total number of effects is $2^p - 1$. This number would become extremely large with the increase of $p$. If all the effects are active or non-zero, then the resulting model can be very complicated and hard to interpret. Fortunately, in tree ensemble models, we can easily control the highest interaction order by maximum tree depth, which is a commonly used hyperparameter. For example, as the maximum tree depth is 1, then all the interaction effects are zero, and the model reduces to a generalized additive model (GAM) with at most $p$ main effects; as the tree depth is 2, then the model would only have main effects and pairwise interactions, which has same model form as the explainable boosting machine (EBM). In this case, the total number of effects is less than or equal to $p(p+1) / 2$. As not all the features are used as split variables, the number of active effects is usually smaller than the number of possible effects.
	
	With a maximum tree depth of 3, we can still interpret the interactions involving 3 features using 3D heatmaps or sliced 1D plots. For example, we can examine a 3-way interaction by visualizing one or two features while keeping the rest one or two features fixed at certain values. However, as the tree depth increases, the model's complexity grows exponentially, making it more challenging to interpret deep tree ensemble models. 
	
	In addition, deep tree ensembles are hard to interpret also from the algorithm perspective. If given adequate computing resources, the purification algorithm can be applied to arbitrary interaction effects. However, the tensors representing high-order interactions tend to become excessively large, making them difficult to process. In practical scenarios, purifying interactions involving 4 or more features becomes challenging, and sometimes even impossible. Hence, to maintain feasibility, our interpretation in this paper is restricted to depth 3 tree ensemble models.
	
	In practice, well-configured shallow tree ensemble models are often sufficient to achieve good predictive performance. It's worth noting that when we limit the maximum depth of base tree learners, it is recommended to increase the number of estimators (boosting rounds). This is because shallow trees in nature have much lower expressive power compared to deeper ones. For instance, a depth 2 tree ensemble model with 100 estimators would have at most 400 leaf nodes, while a similar depth 5 model would have at most 3200 leaf nodes. Therefore, to compensate for the reduction in tree depth, we may need to increase the number of estimators.
	
	\subsubsection{Monotonicity}
	In many real-world applications, enforcing feature monotonicity in a model is highly desirable for interpretation purposes. In a credit scoring model, it is expected that applicants' credit scores increase monotonically with their income. However, in practice, this assumption can be easily violated due to noisy data, rendering the model difficult to interpret and diminishing people's trust in its predictions. In tree ensemble models, monotonicity constraints can be imposed in fitting each tree. For instance, to make a feature monotonic increasing, we can prohibit candidate splits of that feature where the resulting left child node value is greater than that of the right one.
	
	The monotonicity constraint can be specified by leveraging domain knowledge before model training. It can significantly enhance the interpretability and trustworthiness of the model. On the other hand, the EBM model, as a counterpart benchmark, lacks inherent monotonicity constraints, and adjustments can only be made post-training. Such post-hoc adjustments may introduce bias and potentially decrease the overall performance of the model.
	
	\subsubsection{Maximum Number of Bins}
	This parameter is preliminarily employed to reduce the search space of split points. Instead of considering all possible unique feature values as candidate split points, it selects a predetermined number of quantiles for each feature as candidates. From the perspective of model interpretability, restricting the number of bins can also prevent unnecessary discontinuities and make the estimated effects more easily comprehensible. Therefore, this hyperparameter is very useful in practical applications.
	
	\subsubsection{Interaction Constraint}	
	Certain tree ensemble learning frameworks provide an API that allows for the restriction of candidate feature interactions. By using this option, interactions outside of a predefined list of interactions can be prohibited. For example, if we specify the allowed interactions as $(x_1, x_2)$ and $(x_2, x_3)$, the resulting fitted model would not include interactions such as $(x_1, x_3)$. This feature is useful when we possess prior or domain knowledge about the data being modeled, or when we just aim to reduce the complexity of the model.
	
	It is important to note that by applying the feature interaction constraint, the maximum tree depth parameter can be relaxed and set to a larger value without increasing the highest order of interactions. For example, if our goal is to include only main effects and pairwise interactions, we can set the maximum tree depth to a value greater than 2, while constraining the interaction list to encompass all possible pairwise interactions. This approach provides flexibility in hyperparameter tuning, allowing us to vary the depth of the trees while still capturing the desired level of interactions. By using this trick, we can strike a balance between model complexity and interpretability, tailoring the model to our specific requirements.
	
	\subsubsection{Miscellaneous}
	There are several other hyperparameters that can be utilized to enhance the interpretability of tree ensemble models. Here, we outline some of the commonly employed ones:
	
	\begin{itemize}	
		\item \textbf{L1 / L2 Regularization}. Similar to the regularization techniques used in linear models, the application of L1 or L2 regularization can help penalize large values in leaf nodes. By increasing the regularization strength, insignificant leaf nodes can be eliminated, effectively reducing their impact to zero. Consequently, the functional ANOVA representation will also become sparser, making it easier to interpret.
		
		\item \textbf{Early Stopping Conditions}. In addition to the aforementioned criteria, certain early stopping conditions can also be considered as interpretability constraints. These include hyperparameters such as the minimum number of samples per leaf, the minimum loss reduction required for splits, the number of rounds for early stopping, and so on.
	\end{itemize}
	
	\subsection{Effects Pruning}	
	To enhance the interpretability of a tree ensemble model, we can prune trivial effects after it is fitted. This can be approached as a supervised feature selection problem, where each effect is treated as a feature. Various existing feature selection algorithms can be employed to identify the most important effects. In this paper, we introduce two straightforward strategies for effect pruning, as follows.
	
	\subsubsection{Sparse Linear Models}
	A simple approach for effect pruning is to fit a surrogate sparse linear model to identify and remove trivial effects. In this paper, we choose Lasso for regression tasks and L1 regularized logistic regression for classification tasks. The surrogate model can capture the overall relationships between the effects and their impact on the response. It can also identify and flag effects that contribute minimally or have a high correlation with other effects. These effects can be automatically pruned from the model, enhancing its interpretability by focusing on the most relevant and independent effects. 
	
	This pruning strategy shares a similar idea with RuleFit algorithm~\cite{friedman2008predictive}. Both of them try to pursue a parsimonious representation of tree ensemble models by sparse linear modeling. The main difference lies in that RuleFit selects the most important decision rules, while ours perform pruning on the effects functions decomposed by functional ANOVA. Both of these methods are complementary and can also be combined. For instance, one may initially apply pruning on the decision rules level and subsequently represent the selected rules using functional ANOVA. However, such a combination is beyond the scope of this paper.
	
	\subsubsection{Forward or Backward Effect Selection}
	Another powerful strategy is called forward and backward selection with early dropping (FBEDk;~\cite{borboudakis2019forward}). It consists of $k$ forward selection rounds and one backward elimination. The first forward selection round starts from a null model or a pre-defined effect sets and iteratively adds effects that contribute significantly to the model's performance. In the beginning, we regress each effect with the target variable, the effect with the best cross-validation performance will be selected and added to the null model. Also in this step, a filtering step is performed, and all the effects with performance gain greater than the pre-defined threshold are selected as candidate effects. Next, conditional on the selected effect, we select the second one by assessing the performance gain of adding each candidate effect to the linear model. Again, the best one is selected and the candidate effect list is updated. This procedure is repeated until the candidate list is empty. 
	
	As $k > 1$, we would perform multiple rounds of forward selection, and each one starts from the selected effects of previous rounds. Due to the existence of the performance gain threshold, the length of the candidate effects list would become smaller and smaller within each round. Multiple forward rounds are used as it is possible that one effect is not important in the first forward round but will become significant as conditioning on some other effects. Throughout this paper, we set $k=2$.
	
	Finally, as all the forward selection rounds are complete, we do a backward elimination round, starting from the least significant effects. This is testing whether the performance gain of each selected effect (conditioning on the rest selected effects) is greater than the threshold. Effects that fail this test are considered trivial and then removed from the model. As the effects are selected, we refit a generalized linear model between the selected effects and the target variable. The scale of each effect will be changed, while their shape would not. Note that the refitting step may make the model achieve better predictive performance.
	
	\subsubsection{A Hybrid Approach}
	In this paper, we use a hybrid approach that combines the above two strategies.
	\begin{itemize}
		\item First of all, we fit a sparse linear model to roughly select the important effects.
		\item Then, we treat the selected effects as initialization and use the FBEDk algorithm to fine-tune the results. It will access the marginal contribution of each effect, and the ones with contributions greater or less than a pre-defined threshold will be added or deleted accordingly.
		\item Finally, given the selected effects, we refit a linear model without sparsity constraints to adjust the coefficients.
	\end{itemize}
	With this effect pruning algorithm, most of the trivial effects would be removed. The shape of selected effects remains the same, but their scale would be adjusted. In summary, the proposed global effect pruning strategy offers an efficient method for improving the interpretability of the tree ensemble model.

	\section{Numerical Results} \label{experiments}
	In this section, we demonstrate the proposed interpretation algorithm through simulation and real-world datasets. Among the tree ensemble models, we choose the XGB model implemented by the \textsl{xgboost} package throughout the experiments. As maximum depth is the most important hyperparameter, we enumerate maximum depth from 1 to 5, abbreviated as XGB-1, XGB-2, ..., XGB-5. For comparison, the spline-based GAM, EBM, and GAMI-Net are included as benchmark models. The spline-based GAM is implemented by the~\textsl{pyGAM} Python package~\cite{serven2018pygam}. The EBM model is implemented in the Python package~\textsl{interpret}~\cite{nori2019interpretml}. The GAMI-Net model is based on the implementation in the~\textsl{PiML} Python package~\footnote{\url{https://github.com/SelfExplainML/PiML-Toolbox/}}. Moreover, the proposed tree ensemble model interpretation algorithm is also integrated into the \textsl{PiML} package.
	
	We randomly split each dataset into training (80\%) and test (20\%) sets. For hyperparameter tuning and monitoring the early stopping criteria, 20\% of the training samples are used for validation purposes. For each of these 8 XGBoost models, we tune the number of estimators (50 to 3000), learning rate (0.01 to 1), L1 regularization (0.001 to 1000), L2 regularization (0.001 to 1000), and maximum number of bins (2 to 200). In pyGAM, we tune the spline order (0 to 3), number of splines (10 to 50), and smoothing penalty (0.001 to 1000). In EBM, we tune the number of interactions (0 to 100) and the learning rate (0.01 to 1). For each model, we tune the hyperparameters using the random search strategy~\cite{bergstra2012random}, and the number of trials is set to 30. In specific, we randomly generate 30 hyperparameter configurations for each model within the search space; the one that achieves the best validation performance is selected, and then we refit the model using all the training data. For speed consideration, GAMI-Net is configured and trained using the default settings, with the number of interactions fixed to 10.
	
	The predictive performance is measured by the root-mean-square error (RMSE) for regression tasks and the area under the ROC curve (AUC) for binary classification tasks. All the experiments are repeated 10 times, and we report the average results.
	
	\subsection{Simulation Study with Friedman Dataset}
	The Friedman data is based on the following simulation function as described in~\cite{friedman1991multivariate,breiman1996bagging}.
	\begin{equation}
		y(\boldsymbol{x})=10 \sin \left(\pi x_1 x_2\right)+20\left(x_3-0.5\right)^2+10 x_4+5 x_5+\varepsilon,
	\end{equation}
	where $\varepsilon \sim N\left(0, \sigma^2\right)$. The covariates are uniformly distributed between 0 and 1. In this experiment, we simulate data with $n = 2000$ and $\sigma = 0.1$.
	
	\begin{table}[!t]
		\centering
		\renewcommand\tabcolsep{2.5pt}
		\renewcommand\arraystretch{1.1}
		\caption{Comparison results of the Friedman dataset.} \label{results_friedman}
		\begin{tabular}{c|cc|ccc}
			\hline
			& \multicolumn{2}{|c|}{RMSE}        &       \multicolumn{3}{c}{\#Effects}        \\ \hline
			&Train&Test&1-way&2-way&3-way\\ \hline
			XGB-1& 1.251$\pm$0.062 & 1.423$\pm$0.032 & 9.9$\pm$0.3 & 0.0$\pm$0.0 & 0.0$\pm$0.0 \\ 
			XGB-2& 0.223$\pm$0.046 & 0.509$\pm$0.030 & 10.0$\pm$0.0 & 44.9$\pm$0.3 & 0.0$\pm$0.0 \\
			XGB-3& 0.212$\pm$0.062 & 0.571$\pm$0.036 & 10.0$\pm$0.0 & 52.5$\pm$1.1 & 116.9$\pm$3.3 \\
			XGB-4& 0.120$\pm$0.062 & 0.622$\pm$0.035 &-&-&-\\
			XGB-5& 0.143$\pm$0.147 & 0.693$\pm$0.048 &-&-&-\\
			pyGAM& 1.292$\pm$0.038 & 1.359$\pm$0.042 & 10.0$\pm$0.0 & 0.0$\pm$0.0 & 0.0$\pm$0.0 \\
			EBM& 0.289$\pm$0.061 & 0.603$\pm$0.043 & 10.0$\pm$0.0 & 5.2$\pm$2.7 & 0.0$\pm$0.0 \\
			GAMI-Net& 0.122$\pm$0.005 & 0.149$\pm$0.008 & 5.0$\pm$0.0 & 8.7$\pm$1.4 & 0.0$\pm$0.0 \\
			\hline
		\end{tabular}
	\end{table}
	
	Table~\ref{results_friedman} compares the predictive performance of different models on the Friedman dataset. The best model performance of XGB is achieved as the maximum depth is 2, which means the model only captures main effects and pairwise interactions. As the maximum depth further increases, the XGB model tends to overfit the training set, and the test RMSE inversely increases. Compared with EBM, XGB-2 has a much lower RMSE, which means that the latter model is superior to EBM in this specific task. However, we observe that  XGB-1 and XGB-2 are less comparable to their counterpart benchmarks pyGAM and GAMI-Net, respectively. That is because the ground truth function is continuous. pyGAM and GAMI-Net with continuous shape functions can better capture the actual patterns. The tree ensemble models, however, can only approximate the actual patterns via piecewise constant functions. 
	
	\begin{figure}[!t]
		\centering
		\includegraphics[width=0.49\textwidth,height=0.25\textheight]{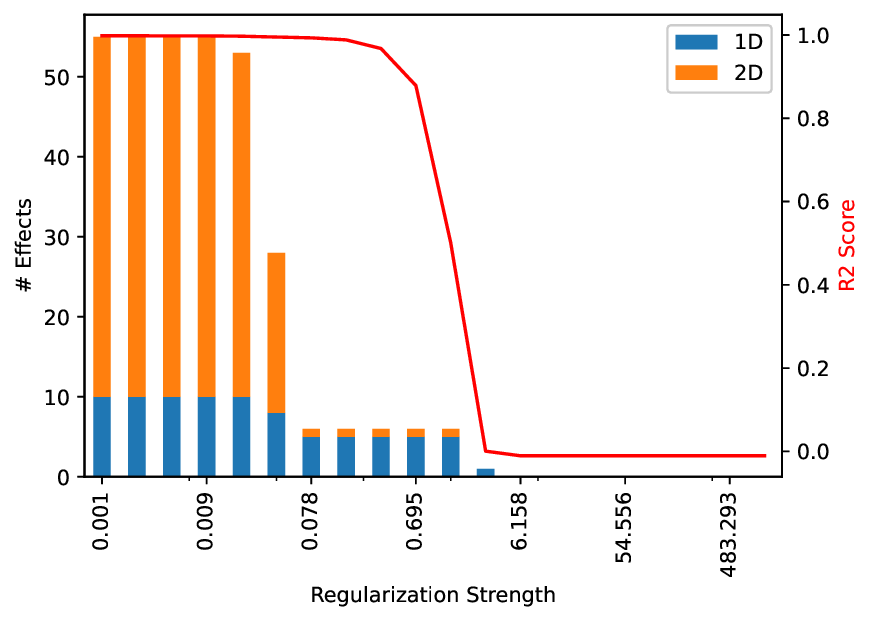} 
		\caption{The number of selected effects and 5-fold cross-validation performance under different regularization strengths of Lasso for the Friedman dataset.}\label{Friedman_regularization_path}	
	\end{figure}
		
	For demonstration, the raw XGB-2 model is interpreted using the proposed algorithm. In total, the raw XGB-2 model is transformed into a functional ANOVA representation with 10 main effects and 45 pairwise interactions. After that, we use Lasso with different regularization strengths to reveal the relationship between predictive performance and the number of selected effects in Figure~\ref{Friedman_regularization_path}. The x-axis is the regularization strength; the bar chart (on the left y-axis) shows the number of selected effects and the line plot (on the right y-axis) displays the 5-fold cross-validation R-squared (R2) score. From the results, it can be observed that R2 reaches its maximum when the regularization is small (from 0.001 to 0.009). As the regularization strength increases to 0.078, the selected effects suddenly shrink to 5 main effects and 1 pairwise interaction, while the R2 score does not change too much. This means that the rest 5 main effects and 44 pairwise interactions are trivial and can be pruned.
				
	\begin{figure}[!t]
		\centering
		\subfloat[Effect Importance]{
			\label{Friedman_effect_importance} 
			\includegraphics[width=0.43\textwidth,height=0.175\textheight]{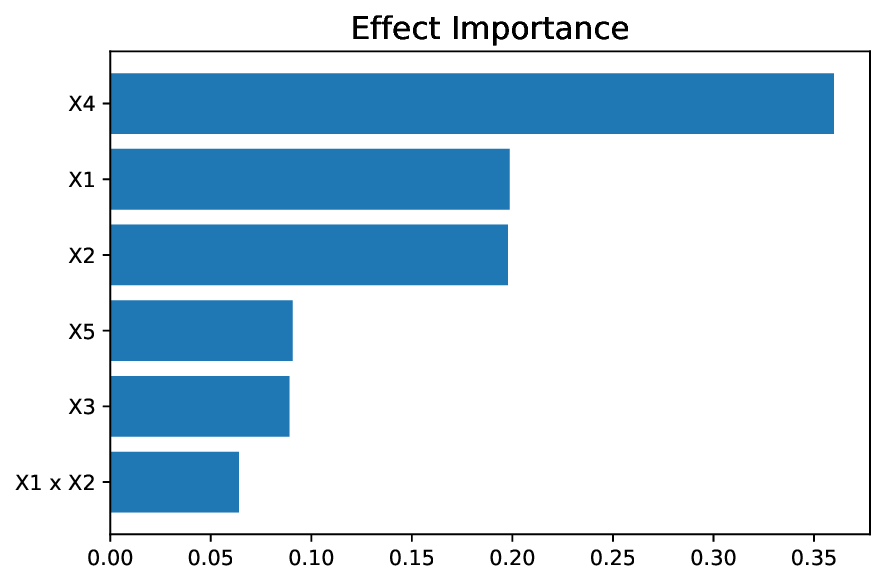}}\\
		\subfloat[Feature Importance]{
			\label{Friedman_feature_importance} 
			\includegraphics[width=0.4\textwidth,height=0.175\textheight]{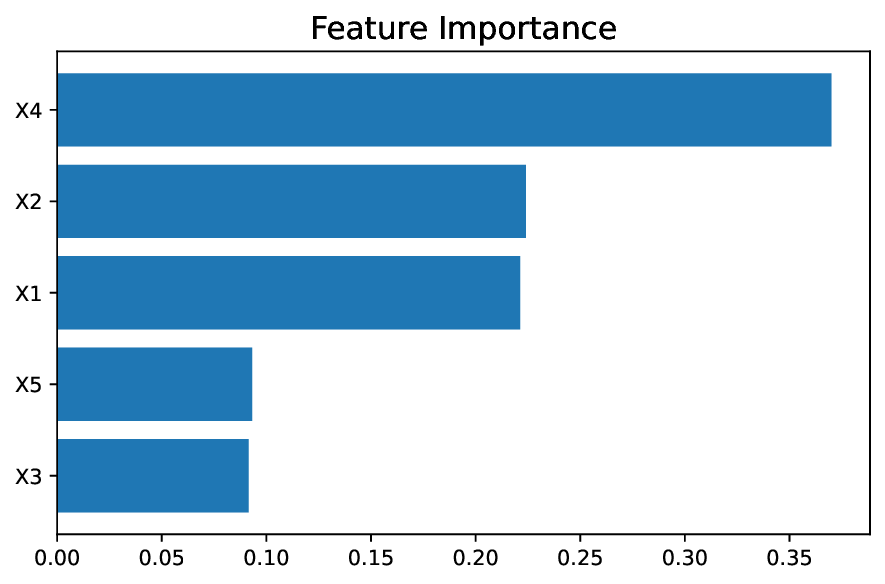}} 
		\caption{The effect and feature importance of the Friedman dataset.}\label{Friedman_importance}
	\end{figure}

	Inspired by this result, we can do post-hoc effect pruning by fitting a Lasso with a regularization strength of 0.078, and then fine-tune the selected effects by the FBEDk algorithm. It is worth mentioning that after effect pruning, the test set RMSE gets improved to around 0.425. This means that removing the trivial effects can not only enhance model interpretability but also mitigate overfitting. Figure~\ref{Friedman_importance} shows the effect and feature importance defined in Section~\ref{methodology}. The 5 main effects $X_1, X_2, \cdots, X_5$ are most important to the model prediction, followed by the interaction $X_1 x X_2$. The feature importance further aggregates the contribution of interactions to each feature. It turns out that $X_4$ is the most important, $X_2, X_1$ are less important, and $X_5, X_3$ are of the least importance.
		
	\begin{figure*}[!t]
		\centering
		\subfloat[Ground Truth]{
			\label{Friedman_GT} 
			\includegraphics[width=0.7\textwidth,height=0.25\textheight]{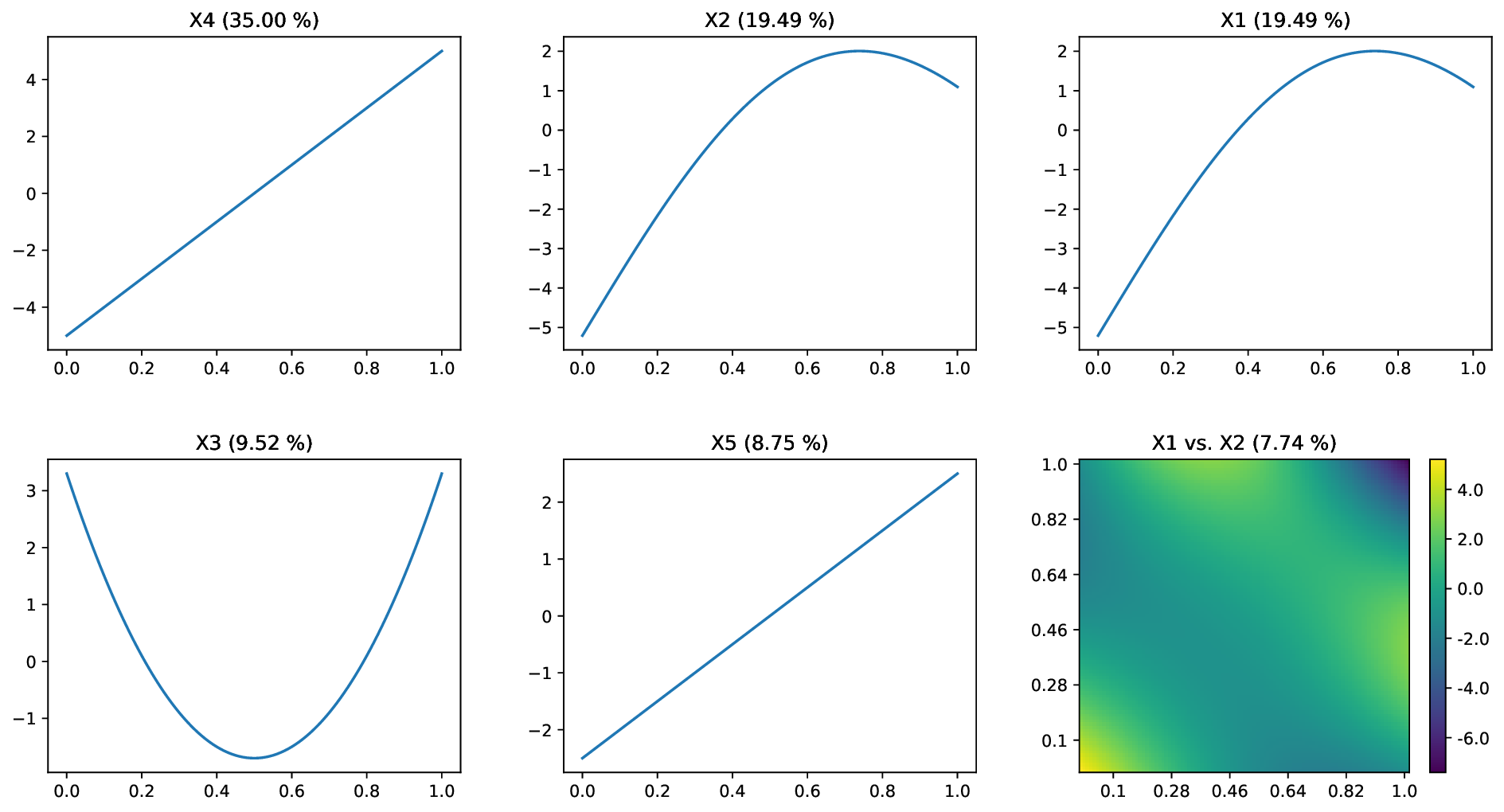}}\\
		\subfloat[XGB-2 with post-hoc effect pruning]{
			\label{Friedman_XGB2} 
			\includegraphics[width=0.7\textwidth,height=0.25\textheight]{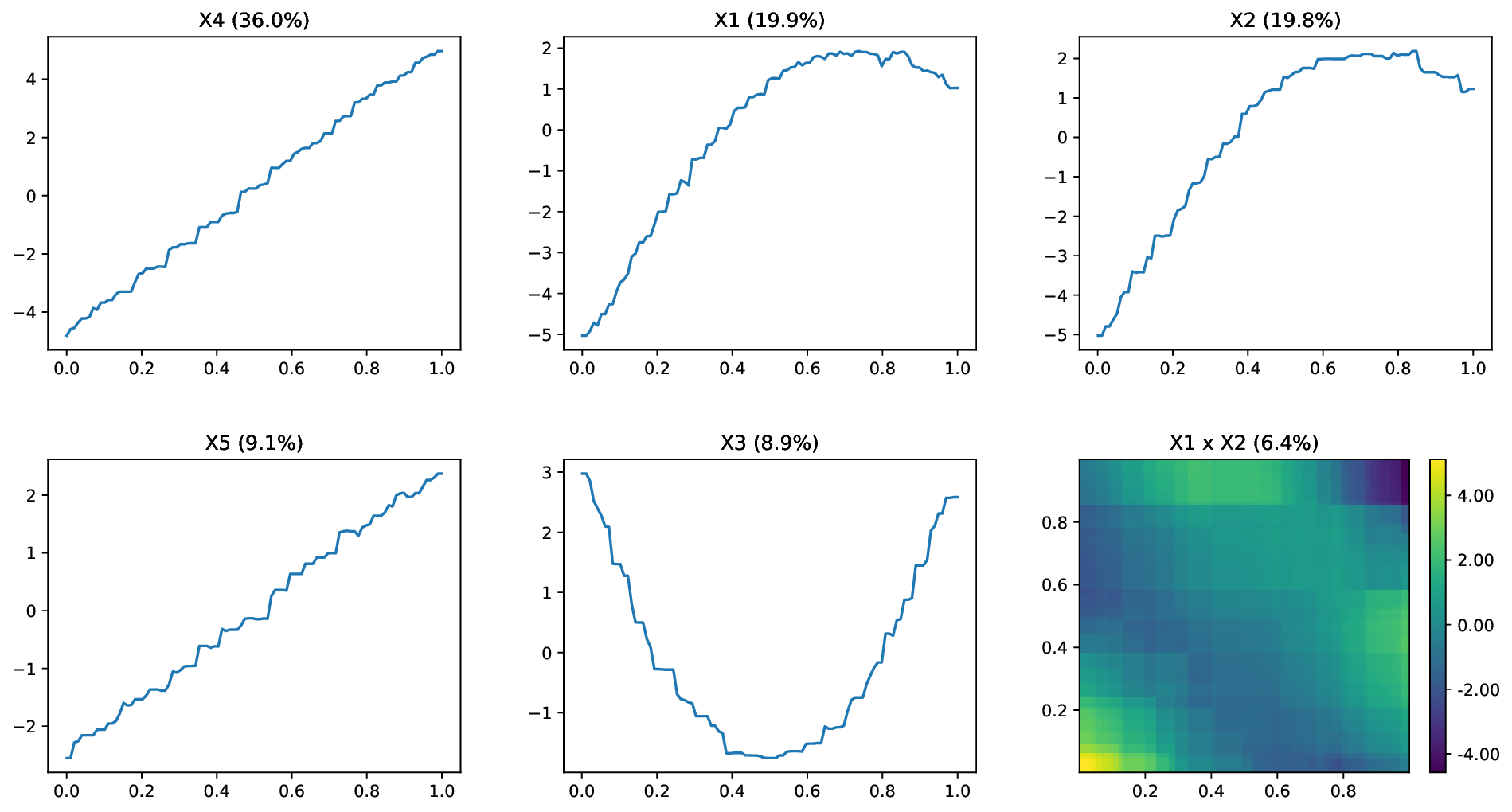}}
		\caption{The fitted results of XGB-2 vs. the ground truth of the Friedman dataset.}\label{Friedman_Visu}	
	\end{figure*}
	
	Figure~\ref{Friedman_Visu} displays the obtained main effects and pairwise interactions after effect pruning, together with the ground truth functions. For each effect plot, we show the corresponding effect importance in the title. Overall, the effects fitted by XGB-2 are close to the actual functions, and the difference is due to the inherent model form of tree ensemble models, i.e., the piecewise constant model fits.
	
	\begin{figure*}[!t]
		\centering
		\subfloat[Local Effect Contribution]{
			\label{Friedman_local_effect} 
			\includegraphics[width=0.49\textwidth,height=0.2\textheight]{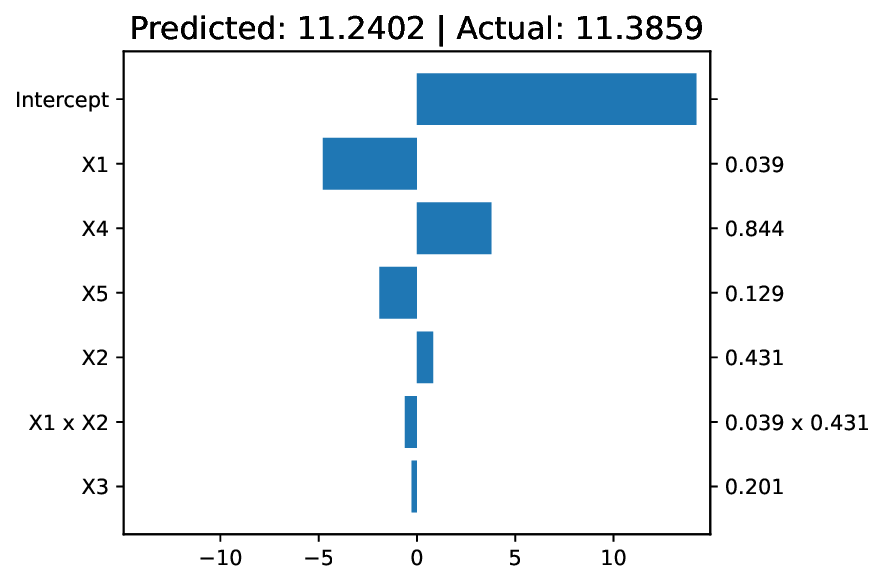}}
		\subfloat[Local Feature Contribution]{
			\label{Friedman_local_feature} 
			\includegraphics[width=0.49\textwidth,height=0.2\textheight]{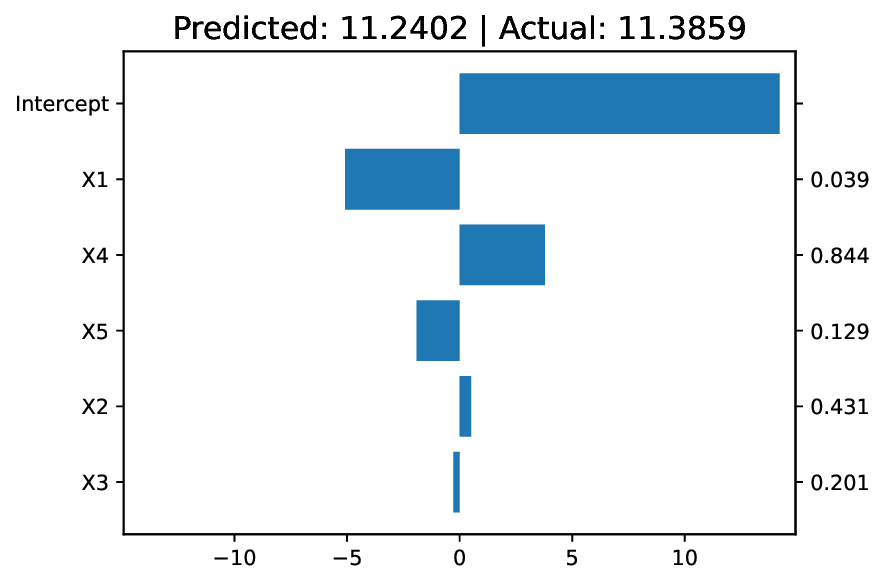}} 
		\caption{A demo of local explanation of the Friedman dataset.}\label{Friedman_local}
	\end{figure*}
	
	Given a specific sample, local explanation tries to explain how the model generates its prediction. The prediction can be additively decomposed into effect contributions and feature contributions, see a demo in Figure~\ref{Friedman_local}. The left axis is the effect / feature names, the right axis shows the feature values of the given sample, and the bar charts represent the contributions of each effect / feature to the prediction. In the title, we also give the predicted value and the actual response.
	
	\subsection{CreditSimu Dataset}
	The second example is a credit decision dataset with synthetic features of applicants, including Mortgage (mortgage size), Balance (average credit card balance), Amount Past Due (minimum required payment that was not applied to the account as of the last payment due date), Delinquency status (0: current, 1: less than 30 days delinquent, 2: 30-60 days delinquent, 3: 60-90 days, etc), Credit Inquiry (number of credit inquiries), Open Trade (number of open credit accounts), and Utilization (credit utilization ratio). This data is provided in the \textsl{PiML} package, and the response feature is binary, indicating whether the application is approved or not.
	
	\begin{table}[!t]
		\centering
		\renewcommand\tabcolsep{4pt}
		\renewcommand\arraystretch{1.1}
		\caption{Comparison results of the CreditSimu dataset.} \label{results_creditsimu}
		\begin{tabular}{c|cc|ccc}
			\hline
			& \multicolumn{2}{|c|}{RMSE}        &       \multicolumn{3}{c}{\#Effects}       \\ \hline
			&Train&Test&1-way&2-way&3-way\\ \hline
			XGB-1& 0.751$\pm$0.002 & 0.745$\pm$0.009 & 7.0$\pm$0.0 & 0.0$\pm$0.0 & 0.0$\pm$0.0 \\
			XGB-2& 0.766$\pm$0.004 & 0.754$\pm$0.008 & 7.0$\pm$0.0 & 18.4$\pm$1.8 & 0.0$\pm$0.0 \\
			XGB-3& 0.770$\pm$0.004 & 0.754$\pm$0.008 & 7.0$\pm$0.0 & 25.5$\pm$1.1 & 27.0$\pm$4.2 \\
			XGB-4& 0.773$\pm$0.006 & 0.754$\pm$0.007 &-&-&-\\
			XGB-5& 0.774$\pm$0.005 & 0.753$\pm$0.008 &-&-&-\\
			pyGAM& 0.743$\pm$0.021 & 0.740$\pm$0.022 & 7.0$\pm$0.0 & 0.0$\pm$0.0 & 0.0$\pm$0.0 \\
			EBM& 0.768$\pm$0.002 & 0.753$\pm$0.007 & 7.0$\pm$0.0 & 10.5$\pm$6.3 & 0.0$\pm$0.0 \\
			GAMI-Net& 0.756$\pm$0.003 & 0.753$\pm$0.008 & 4.2$\pm$0.4 & 1.7$\pm$0.7 & 0.0$\pm$0.0 \\
			\hline
		\end{tabular}
	\end{table}
	
	The comparison results are shown in Table~\ref{results_creditsimu}. From the results, we can find that the XGB-2 model reaches the best test set AUC, and adding higher-order interactions will not increase its predictive performance. The EBM and GAMI-Net models also have very close predictive performance. All the models with at least 2-way interactions outperform the ones with only main effects, i.e., XGB-1 and pyGAM. This reveals the necessity of including pairwise interactions in this task.
	
	According to our domain knowledge, it is expected that some of the features are monotonic with respect to the credit card approval rate. In addition to the raw XGB-2 model, we fit another XGB-2 model with enhanced interpretability constraints. In specific, we constrain the Mortgage to be monotonically increasing and Utilization to be monotonically decreasing. We also limit the maximum number of bins to 20 to avoid unnecessary jumps in fitted shape functions. The constrained XGB-2 model still achieves a test AUC score of around 0.754. This means that the interpretability constraint does not come with any sacrifice in predictive performance.
	
	\begin{figure}[!t]
		\centering
		\includegraphics[width=0.49\textwidth,height=0.25\textheight]{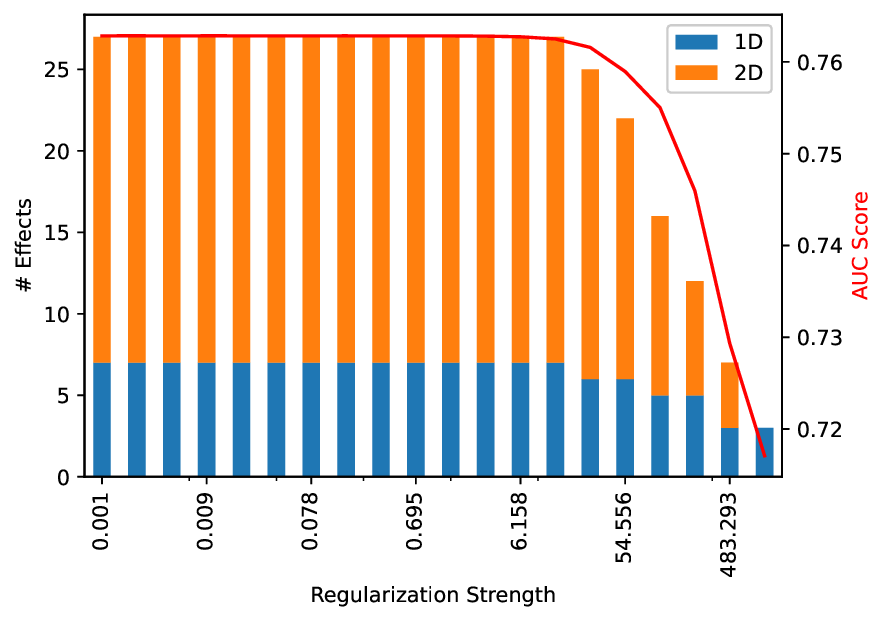} 
		\caption{The number of selected effects and 5-fold cross-validation performance under different regularization strengths of Lasso for the CreditSimu dataset.}\label{CreditSimu_regularization_path}	
	\end{figure}
	
	\begin{figure*}[!t]
		\centering
		\subfloat[Raw XGB-2]{
			\label{CreditSimu_raw} 
			\includegraphics[width=0.9\textwidth,height=0.13\textheight]{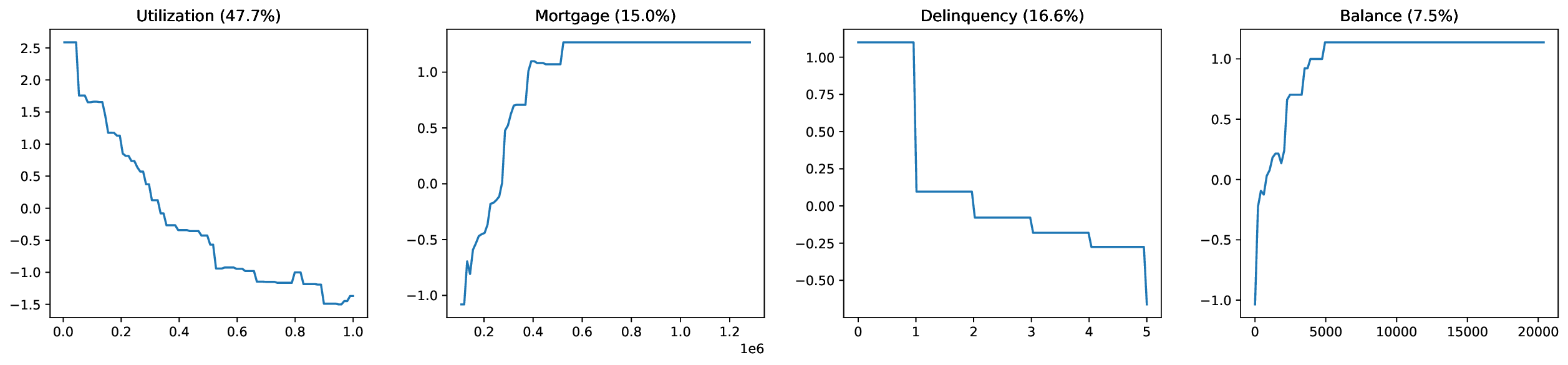}}\\
		\subfloat[Constrained XGB-2]{
			\label{CreditSimu_constrained} 
			\includegraphics[width=0.9\textwidth,height=0.13\textheight]{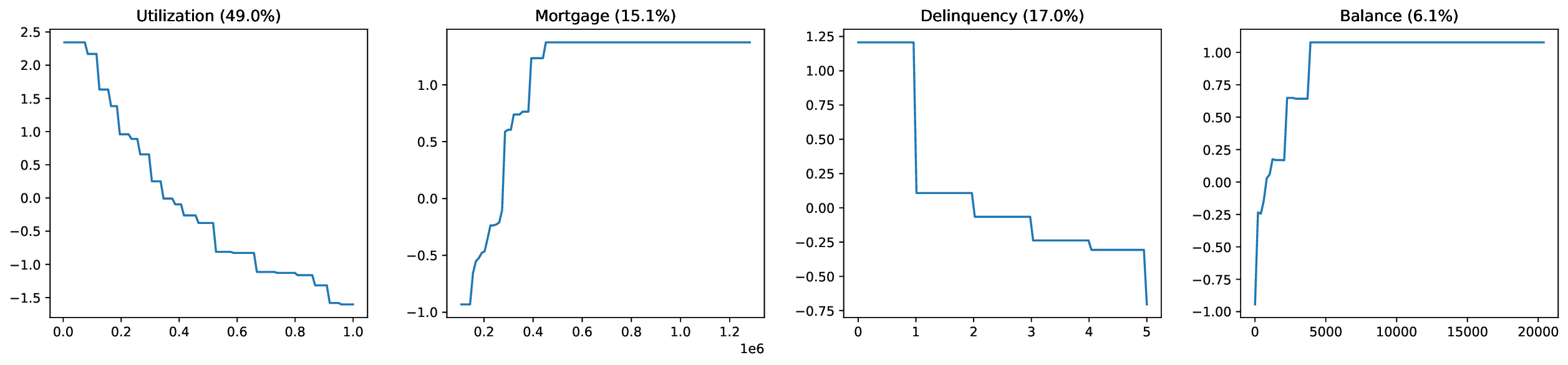}}\\
		\subfloat[Constrained XGB-2 with post-hoc effect pruning]{
			\label{CreditSimu_simplified} 
			\includegraphics[width=0.9\textwidth,height=0.39\textheight]{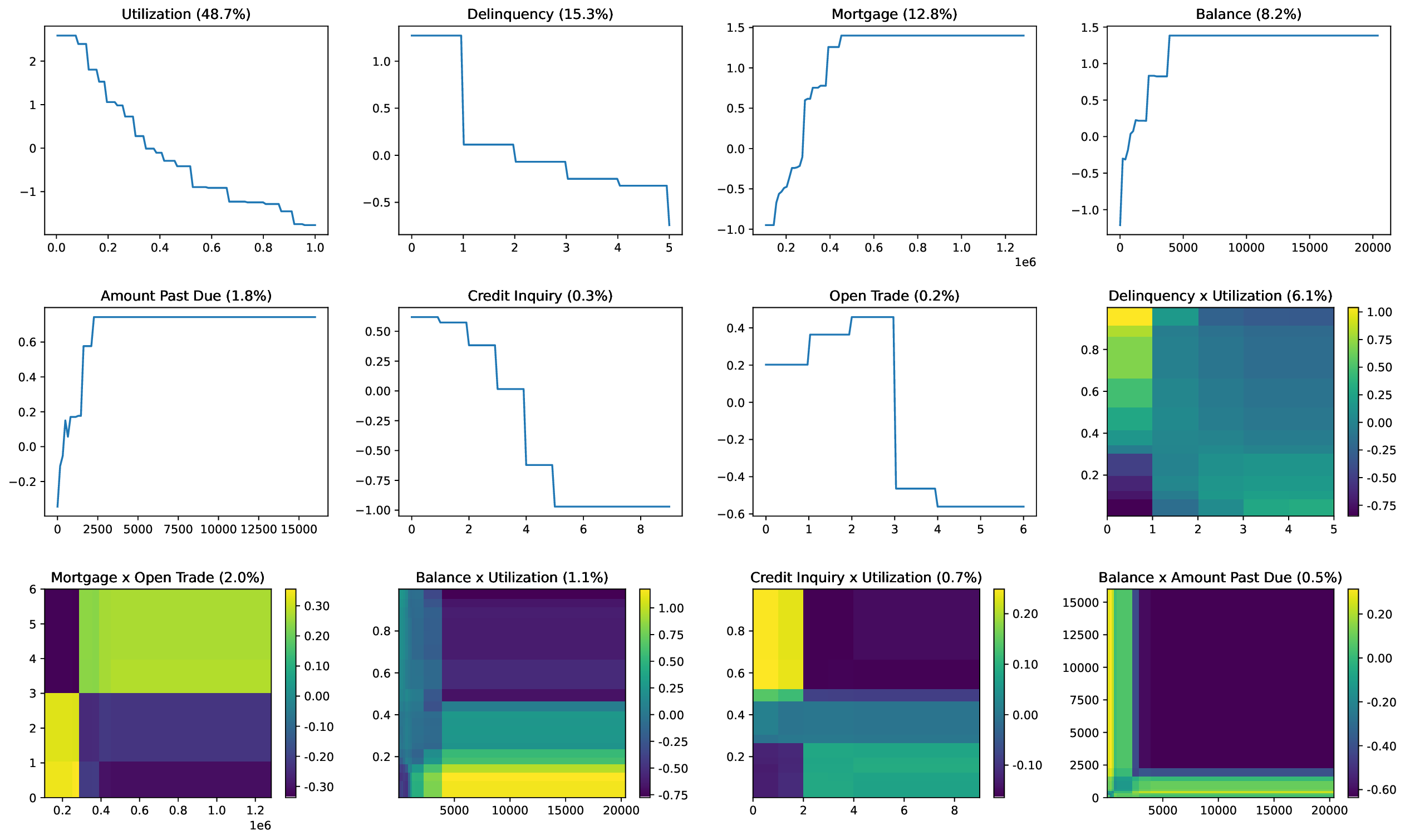}}
		\caption{The visualization comparison of different versions of XGB-2 on the CreditSimu dataset.}\label{CreditSimu_Visu}	
	\end{figure*}
	
	Finally, we also show the Lasso regulation path of the constrained XGB-2 upon functional ANOVA decomposition in Figure~\ref{CreditSimu_regularization_path}. According to the trade-off between model sparsity and AUC score, we prune the constrained XGB-2 via L1-regularization logistic regression (with regularization strength equals 10) and FBEDk (for fine-tuning). The pruned model has a test AUC score of around 0.753 but only includes 7 main effects and 15 pairwise interactions. In Figure~\ref{CreditSimu_Visu}, we show the extracted main effects and pairwise interactions of the pruned XGB-2 model. Note that this is a binary classification task, and hence the y-axis is the log odds ratio. For comparison purposes, we also display the effects of the raw XGB-2 and constrained XGB-2. For simplification, we only show the effects of Mortgage, Utilization, Delinquency, and Balance. It can be found that these constraints make the shape functions of main effects less jumpy and easier to interpret.
	
	\subsection{BikeSharing Dataset}
	This dataset records the hourly count of rental bikes in the Capital bikeshare system from 2011 to 2012. It has 17389 samples, each data record captures the weather and seasonal conditions within an hour, and the task is to predict the total rental bikes including both casual and registered. For modeling purposes, we remove some of the highly redundant variables. The selected predictors include season, hr (hour of a day), holiday (weather day is holiday or not), weekday (day of the week), weathersit (weather conditions), atemp (normalized feeling temperature), hum (normalized humidity), and windspeed (normalized wind speed). As the response is counting data, we process it using log transformation.
	
	\begin{table}[!t]
		\centering
		\renewcommand\tabcolsep{3pt}
		\renewcommand\arraystretch{1.1}
		\caption{Comparison results of the BikeSharing dataset.} \label{results_bike_share}
		\begin{tabular}{c|cc|ccc}
			\hline
			& \multicolumn{2}{|c|}{RMSE}        &       \multicolumn{3}{c}{\#Effects}       \\ \hline
			&      Train      &      Test       &    1-way    &    2-way     &    3-way     \\ \hline
			XGB-1   & 0.650$\pm$0.002 & 0.662$\pm$0.008 & 8.0$\pm$0.0 & 0.0$\pm$0.0  & 0.0$\pm$0.0  \\
			XGB-2   & 0.378$\pm$0.006 & 0.413$\pm$0.008 & 8.0$\pm$0.0 & 28.0$\pm$0.0 & 0.0$\pm$0.0  \\
			XGB-3   & 0.343$\pm$0.013 & 0.401$\pm$0.010 & 8.0$\pm$0.0 & 38.8$\pm$0.4 & 54.4$\pm$1.0 \\
			XGB-4   & 0.310$\pm$0.025 & 0.396$\pm$0.008 &      -      &      -       &      -       \\
			XGB-5   & 0.292$\pm$0.031 & 0.395$\pm$0.008 &      -      &      -       &      -       \\
			pyGAM   & 0.650$\pm$0.002 & 0.662$\pm$0.008 & 8.0$\pm$0.0 & 0.0$\pm$0.0  & 0.0$\pm$0.0  \\
			EBM    & 0.376$\pm$0.009 & 0.417$\pm$0.010 & 8.0$\pm$0.0 & 19.9$\pm$7.2 & 0.0$\pm$0.0  \\
			GAMI-Net & 0.429$\pm$0.008 & 0.439$\pm$0.014 & 5.5$\pm$0.7 & 8.6$\pm$1.1  & 0.0$\pm$0.0  \\ \hline
		\end{tabular}
	\end{table}
	
	In Table~\ref{results_bike_share}, the comparison results of the Bike Sharing dataset are shown. As this is a real-world application, the actual functional form and the highest interaction order are unknown. In this case, we observe that the XGB-5 model achieves the best test set RMSE, and it is possible that deeper XGB may perform even better. However, the performance gain by increasing tree depth seems to be small; the XGB-3 model also has quite good RMSE. To pursue better interpretability, it is worthy of slightly sacrificing the predictive performance, and we select XGB-3 as the target model to be explained.
	
	\begin{figure}[!t]
		\centering
		\includegraphics[width=0.49\textwidth,height=0.25\textheight]{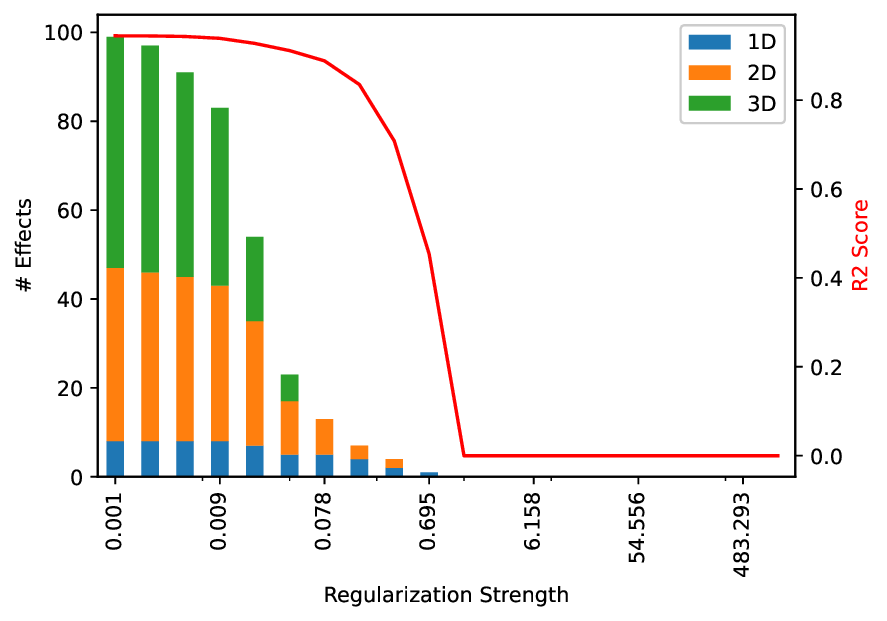} 
		\caption{The number of selected effects and 5-fold cross-validation performance under different regularization strengths of Lasso for the BikeSharing dataset.}\label{BikeSharing_regularization_path}	
	\end{figure}
	
	To make the XGB-3 model more interpretable, we analyze the relationship between the number of effects and the predictive performance using the Lasso regularization path plot, as shown in Figure~\ref{BikeSharing_regularization_path}. Based on this plot, we conduct post-hoc effect pruning with Lasso (regularization strength equals 0.005) and FBEDk (for fine-tuning). Finally, the pruned model has 8 main effects, 37 pairwise interactions, and 46 3-way interactions. The pruned model has a test set RMSE of around 0.408, which is close to that of the raw XGB-3 model (0.406, with the same random seed).
	
	\begin{figure*}[!t]
		\centering
		\subfloat[Raw XGB-3]{
			\label{BikeSharing_raw} 
			\includegraphics[width=0.9\textwidth,height=0.13\textheight]{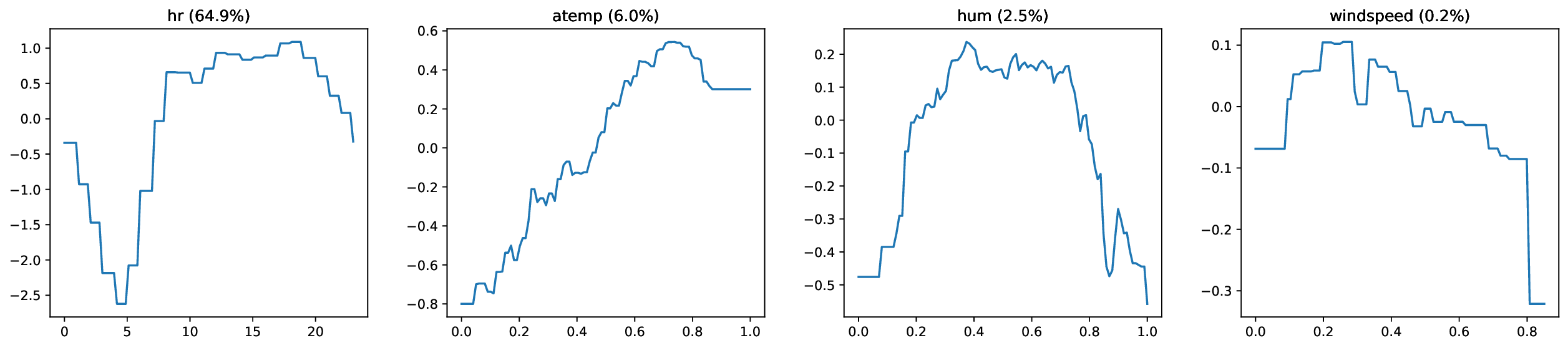}}\\
		\subfloat[XGB-3 with post-hoc effect pruning]{
			\label{BikeSharing_simplified} 
			\includegraphics[width=0.9\textwidth,height=0.39\textheight]{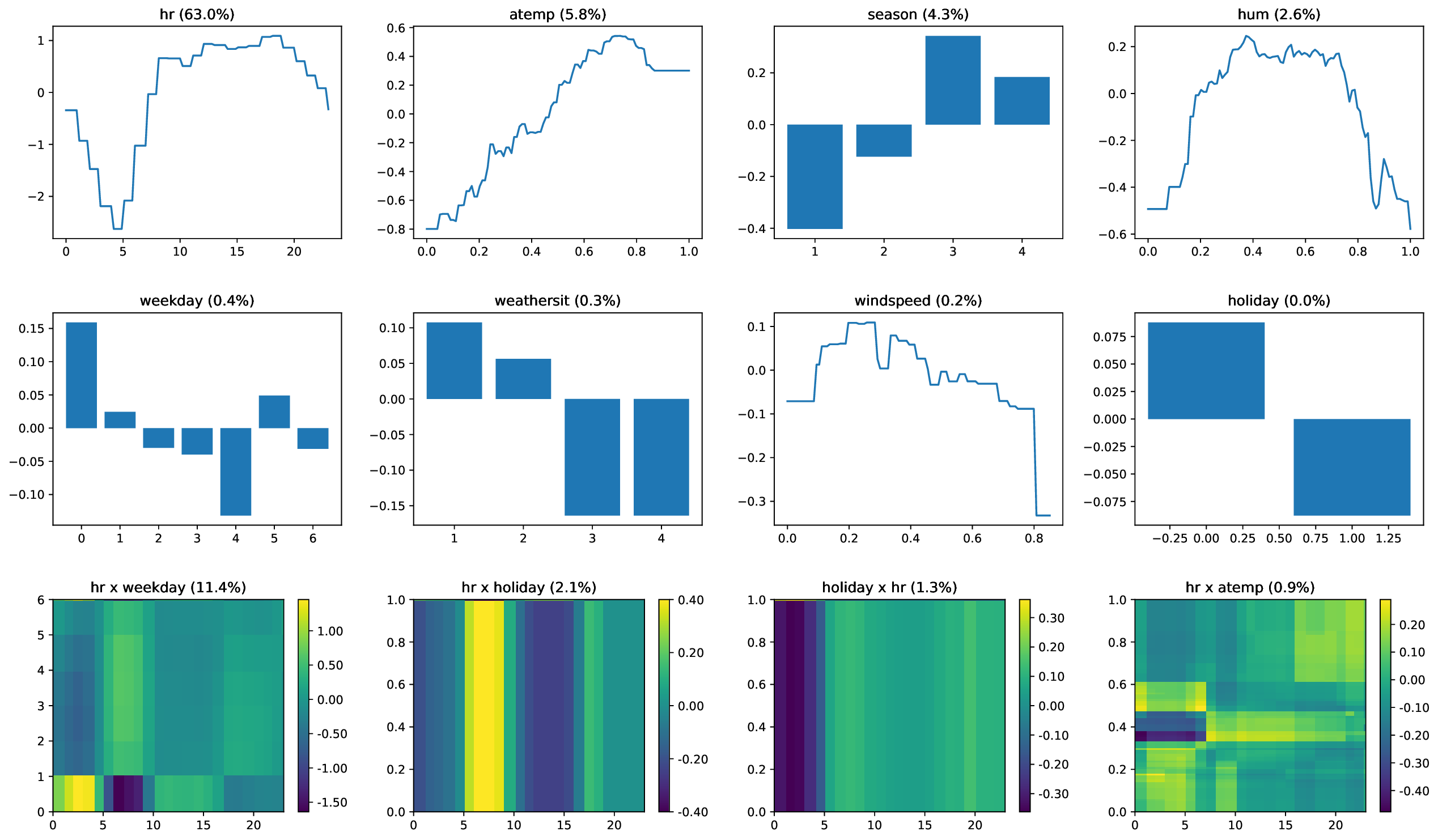}}
		\caption{The visualization comparison of different versions of XGB-2 on the BikeSharing dataset.}\label{BikeSharing_Visu}	
	\end{figure*}
	
	\begin{figure*}[!t]
		\centering
		\includegraphics[width=0.7\textwidth,height=0.25\textheight]{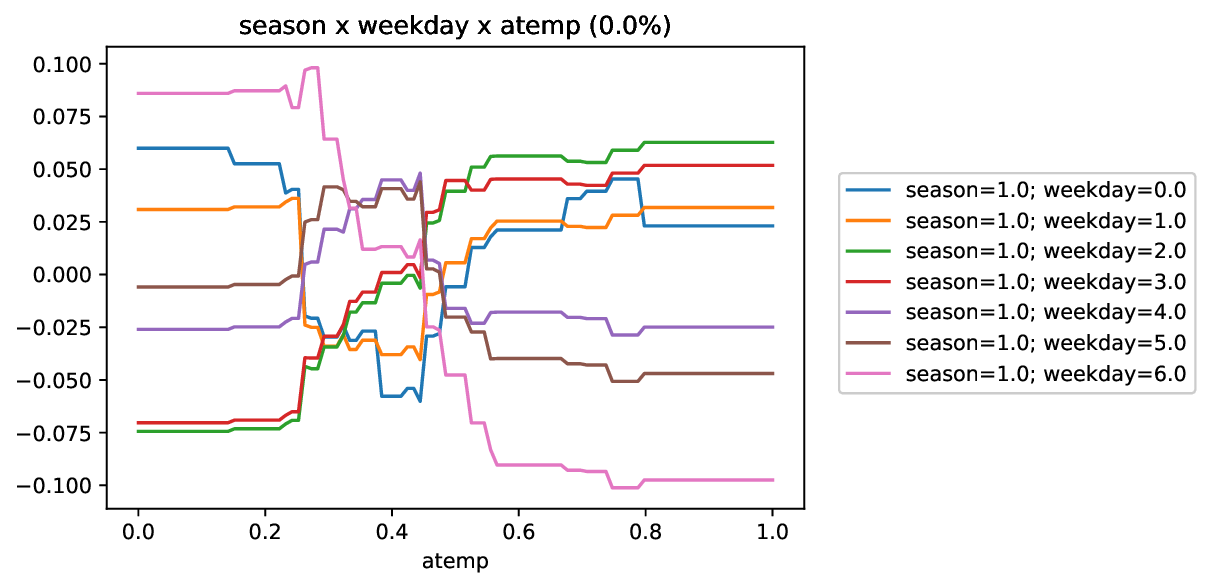} 
		\caption{Visualization of a 3-way interaction of the BikeSharing dataset.}\label{BikeSharing_effect_plots_3D}	
	\end{figure*}
	
	Figure~\ref{BikeSharing_Visu} displays the most important main effects and pairwise interactions in this XGB-3 model, compared with the top main effects of the raw XGB-3. For 3-way interactions, we can use the sliced 1D plot to reveal the patterns. For example, the most important 3-way interaction is season, weekday, and atemp. Conditioning on different value combinations of season and weekday, we draw this interaction value against atamp in Figure~\ref{BikeSharing_effect_plots_3D}. This plot only uses season=1.0, and the other values of season can also be drawn in other plots. It reveals that atemp has an increasing trend to the target as season=1.0 and weekday is 2.0 or 3.0 (Tuesday or Wednesday); however, a decreasing trend is observed as weekday is 6.0 (Saturday).

	\subsection{TaiwanCredit Dataset}
	TaiwanCredit data is obtained from the UCI repository, which consists of 30,000 credit card clients in Taiwan from 200504 to 200509. In this experiment, we only use the 18 payment feature as predictors, including Pay\_1 to 6 (past payment delay status), BILL\_AMT1 to 6 (amount of bill statement), and PAY\_AMT1 to 6 (amount of previous payment). Note that Pay\_1 is renamed from Pay\_0 in the original data. The target variable is default payment, with 1 indicating default payment.
	
	\begin{table}[!t]
		\centering
		\renewcommand\tabcolsep{1.5pt}
		\renewcommand\arraystretch{1.1}
		\caption{Comparison results of the TaiwanCredit dataset.} \label{results_taiwancredit}
		\begin{tabular}{c|cc|ccc}
			\hline
			& \multicolumn{2}{|c|}{RMSE}        &          \multicolumn{3}{c}{\#Effects}          \\ \hline
			&Train&Test&1-way&2-way&3-way\\ \hline
			XGB-1& 0.781$\pm$0.003 & 0.772$\pm$0.007 & 17.7$\pm$0.7 & 0.0$\pm$0.0 & 0.0$\pm$0.0 \\
			XGB-2& 0.787$\pm$0.008 & 0.774$\pm$0.006 & 18.0$\pm$0.0 & 93.3$\pm$27.5 & 0.0$\pm$0.0 \\
			XGB-3& 0.794$\pm$0.012 & 0.776$\pm$0.007 & 18.0$\pm$0.0 & 159.8$\pm$28.5 & 265.0$\pm$115.9 \\
			XGB-4& 0.795$\pm$0.007 & 0.776$\pm$0.006 &-&-&-\\
			XGB-5& 0.797$\pm$0.014 & 0.775$\pm$0.006 &-&-&-\\
			pyGAM& 0.781$\pm$0.002 & 0.773$\pm$0.007 & 18.0$\pm$0.0 & 0.0$\pm$0.0 & 0.0$\pm$0.0 \\
			EBM& 0.793$\pm$0.002 & 0.774$\pm$0.007 & 18.0$\pm$0.0 & 48.9$\pm$22.0 & 0.0$\pm$0.0 \\
			GAMI-Net& 0.775$\pm$0.002 & 0.770$\pm$0.007 & 12.1$\pm$2.2 & 1.2$\pm$0.8 & 0.0$\pm$0.0 \\
			\hline
		\end{tabular}
	\end{table}
	
	Table~\ref{results_taiwancredit} summarizes the AUC comparison of different models on the TaiwanCredit dataset. The best test set performance is obtained by XGB-3. However, XGB-1 and XGB-2 also have very good AUC scores, i.e., 0.772 and 0.774, respectively. The rest compared models all have similar performance, with test AUC ranging from 0.770 to 0.776.
	
	\begin{figure}[!t]
		\centering
		\includegraphics[width=0.49\textwidth,height=0.25\textheight]{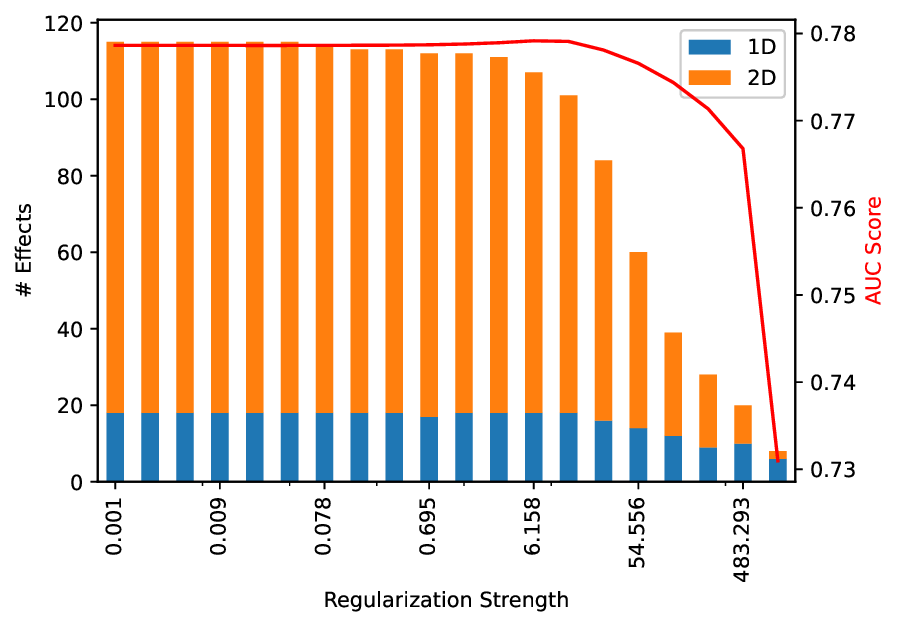} 
		\caption{The number of selected effects and 5-fold cross-validation performance under different regularization strengths of Lasso for the TaiwanCredit dataset.}\label{TaiwanCredit_regularization_path}	
	\end{figure}
	
	Similar to the previous analysis, we first add monotonically increasing constraints for Pay\_1 to 6 (history of past payment) and BILL\_AMT1 to 6; while PAY\_AMT1 to 6 are enforced to be monotonically decreasing. In addition, the maximum number of bins is set to 20. With these interpretability constraints, the constrained XGB-2 model has a test set AUC score of around 0.772, which is slightly lower than that of the raw XGB-2 (around 0.774). The regularization path of the constrained XGB-2 model is drawn in Figure~\ref{TaiwanCredit_regularization_path}. Using this plot, we set the regularization strength of L1-regularized logistic regression to 15 and fine-tune the selected effects by FBEDk. The final pruned model has 15 main effects and 24 pairwise interactions, which is much smaller than the non-pruned model (18 main effects and 97 pairwise interactions). Meanwhile, the pruned model also has a test set AUC score of around 0.772.
	
	\begin{figure*}[!t]
		\centering
		\subfloat[Raw XGB-2]{
			\label{TaiwanCredit_raw} 
			\includegraphics[width=0.9\textwidth,height=0.13\textheight]{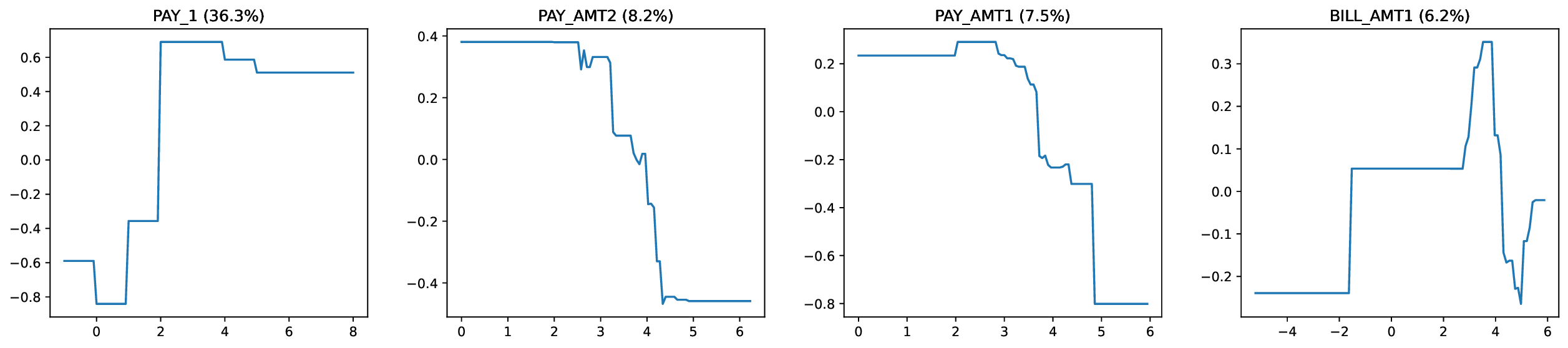}}\\
		\subfloat[Constrained XGB-2]{
			\label{TaiwanCredit_constrained} 
			\includegraphics[width=0.9\textwidth,height=0.13\textheight]{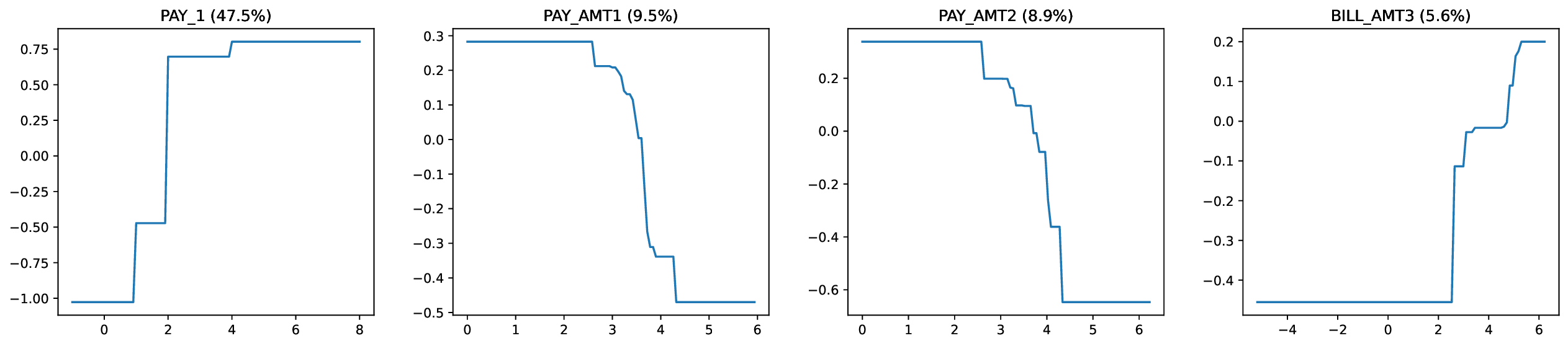}}\\
		\subfloat[Constrained XGB-2 with post-hoc effect pruning]{
			\label{TaiwanCredit_simplified} 
			\includegraphics[width=0.9\textwidth,height=0.26\textheight]{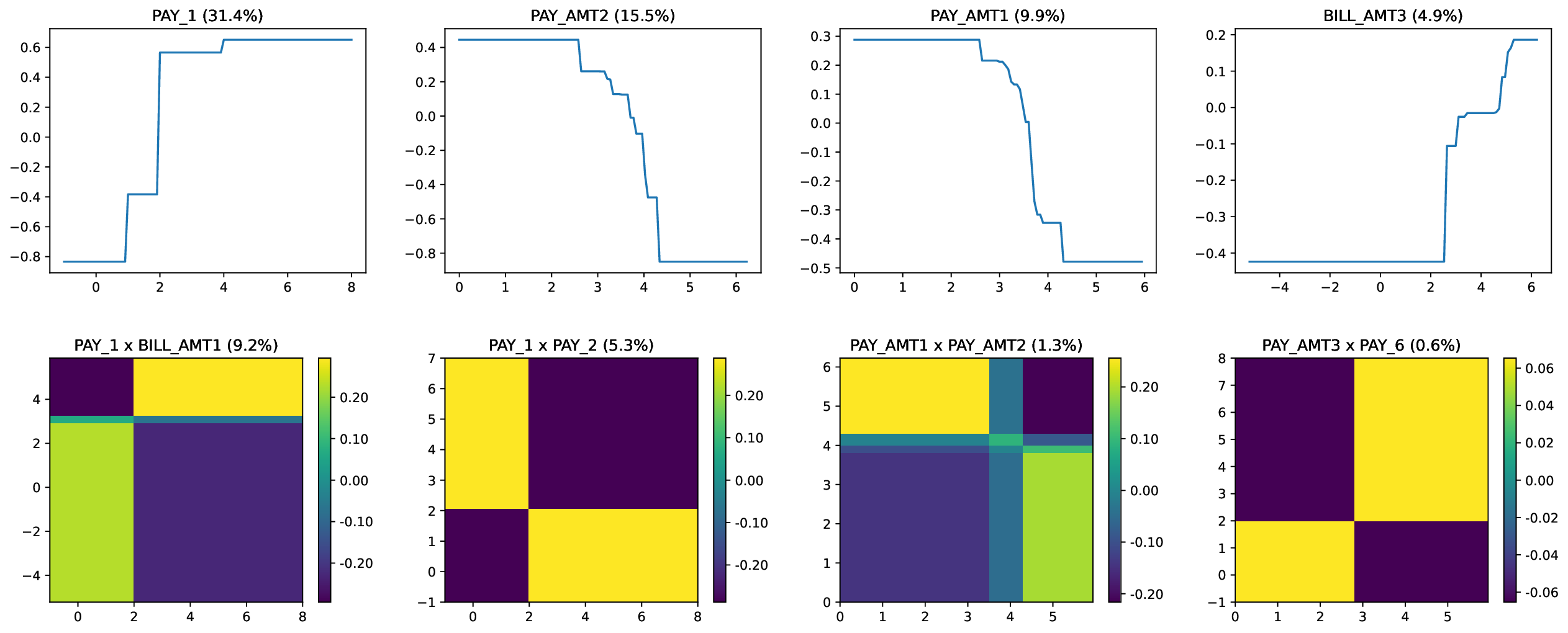}}
		\caption{The visualization comparison of different versions of XGB-2 on the TaiwanCredit dataset.}\label{TaiwanCredit_Visu}	
	\end{figure*}
	
	Figure~\ref{TaiwanCredit_Visu} shows the effects visualization of the 3 versions of XGB-2 models. Clearly, with enhanced interpretability constraints, the effect functions look much more reasonable as compared to the unconstrained one. This further verified our belief that properly imposed constraints can make machine learning models more interpretable.

	\section{Conclusion} \label{conclusion}
	This paper proposes an interpretation algorithm to open the black box of tree ensemble models. Based on the functional ANOVA framework, a fitted tree ensemble model can be equivalently converted into the generalized additive model with interactions.
	Each of the decomposed main effects and pairwise interactions can be easily interpreted and visualized. High-way interactions are more difficult to interpret, however, we empirically show that they are less important and sometimes can be pruned without sacrificing too much predictive performance.
	
	The proposed algorithm is simple but efficient in explaining the fitted tree ensemble models. It can be used for arbitrary tree ensemble models. The following topics are worth investigating. First, although current tree ensemble implementations support monotonicity constraints, they still lack other shape constraints, e.g., convex, concave, etc. Second, it is a promising direction to develop better visualization tools to display high-way interactions.

	\bibliographystyle{IEEEtran}
	\bibliography{IEEEabrv,Manuscript_IEEE}

\end{document}